\documentclass[lettersize,journal]{IEEEtran}
\usepackage{amsmath,amsfonts}
\usepackage{algorithmic}
\usepackage{algorithm}
\usepackage{array}
\usepackage[caption=false, font=footnotesize]{subfig}
\usepackage{textcomp}
\usepackage{stfloats}
\usepackage{url}
\usepackage{verbatim}
\usepackage{graphicx}
\usepackage{cite}
\usepackage{siunitx}
\usepackage{amsmath}
\usepackage{subcaption}
\begin{document}

\title{Federated Learning framework for LoRaWAN-enabled IIoT communication: A case study}

\author{Oscar Torres Sanchez, Guilherme Borges, Duarte Raposo, André Rodrigues, Fernando Boavida, Jorge Sá Silva
\thanks{ This work was supported by FCT – Foundation for Science and Technology by supporting a Ph.D. grant with reference 2023.02113.BD; also within the scope of CISUC - UID/CEC/00326/2020 and European Social Fund, through the Regional Operational Program Centro 2020; and lastly, by Portuguese Foundation for Science and Technology under the project grant UIDB/00308/2020 with the DOI 10.54499/UIDB/00308/2020 and the project 2023 - PDS \& VDS from INESC Coimbra.}

\thanks{O. Torres Sanchez and F. Boavida are with Centre of Informatics and Systems of the University of Coimbra, CISUC, 3030-290, Coimbra, Portugal (email: otorres@dei.uc.pt, boavida@dei.uc.pt).}

\thanks{G. Borges and J. Sá Silva, are with University of Coimbra, Institute for Systems Engineering and Computers, INESC, 3030-790, Coimbra, Portugal and G. Borges is also with Sul-Rio-Grandense Federal Institute (IFSUL), Campus Charqueadas, 96745-000, Brazil (email: guilhermeborges@ifsul.edu.br, sasilva@deec.uc.pt)}
\thanks{D. Raposo is with Instituto de Telecomunicações, 3810-193, Aveiro, Portugal (email: dmgraposo@av.it.pt).} 
\thanks{André Rodrigues is with with Polytechnic Institute of Coimbra, Coimbra Business School, Quinta Agrícola - Bencanta, 3045-231 Coimbra, Portugal. INESC Coimbra – DEEC, University of Coimbra, Polo 2, 3030-290, Coimbra, Portugal. CEOS.PP Coimbra, Polytechnic Institute of Coimbra, Bencanta, 3045-601 Coimbra, Portugal (email: andre@iscac.pt)}
\thanks{Manuscript received April 19, 2021; revised August 16, 2021.}}

\markboth{O. Torres Sanchez \MakeLowercase{\textit{et al.}} :Federated Learning framework for LoRaWAN-enabled IIoT communication: A case study }
{Shell \MakeLowercase{\textit{et al.}}: A Sample Article Using IEEEtran.cls for IEEE Journals}


\maketitle

\begin{abstract}

The development of intelligent Industrial Internet of Things (IIoT) systems promises to revolutionize operational and maintenance practices, driving improvements in operational efficiency. Anomaly detection within IIoT architectures plays a crucial role in preventive maintenance and spotting irregularities in industrial components. However, due to limited message and processing capacity, traditional Machine Learning (ML) faces challenges in deploying anomaly detection models in resource-constrained environments like LoRaWAN. On the other hand, Federated Learning (FL) solves this problem by enabling distributed model training, addressing privacy concerns, and minimizing data transmission. This study explores using FL for anomaly detection in industrial and civil construction machinery architectures that use IIoT prototypes with LoRaWAN communication. The process leverages an optimized autoencoder neural network structure and compares federated models with centralized ones. Despite uneven data distribution among machine clients, FL demonstrates effectiveness, with a mean F1 score (of 94.77), accuracy (of 92.30), TNR (of 90.65), and TPR (92.93), comparable to centralized models, considering airtime of trainning messages of 52.8 min. Local model evaluations on each machine highlight adaptability. At the same time, the performed analysis identifies message requirements, minimum training hours, and optimal round/epoch configurations for FL in LoRaWAN, guiding future implementations in constrained industrial environments.
\end{abstract}

\section{Introduction}

The industry has seen a growing drive for productivity in recent years as it implements increasingly intelligent and autonomous processes based on machine learning (ML) models. The current machine learning paradigm for industrial environments involves large and complex models that may improve product, machine, or system processes \cite{Jan2023}. These models, therefore, operate with abundant resources, training large amounts of data, using high-capacity processing servers and high-bandwidth communication networks such as LAN, 5G, and Wi-Fi, among others. However, in many cases, industrial ML architectures are driven by Industrial Internet of Things (IIoT) devices, operating in industries with limited bandwidth networks, such as low-power WAN (LP-WAN) \cite{Abadade2023}, which require adaptations to the paradigm. Industries such as manufacturing \cite{Wan2021}, mining \cite{Pouresmaieli2023}, renewable energy \cite{Charef2023}, construction \cite{Statsenko2023}, transportation \cite{Gong2023}, logistics, and agriculture \cite{Mowla2023} can benefit from adapting ML models powered by IIoT devices using LoRaWAN to provide cost-effective connectivity solutions that enable real-time data collection, analysis, and decision-making. However, it is important to recognize the inherent limitations of IIoT devices based on LP-WAN networks, which train and run ML models with low processing capacity, using private and sensitive industry data \cite{Peter2023}. These characteristics demand the exploration of alternative ML methods tailored to these limitations.

This study looks at anomaly detection in industrial data,  gathered on constrained LP-WAN scenarios. Anomaly detection facilitated by ML, such as autoencoder-based models, enables the identification of deviations from normal machine behavior by leveraging essential sensor data. While the literature predominantly revolves around cloud-based ML architectures driven by data collected over extended periods, such approaches have limitations in IIoT environments\cite{Rodriguez2023, Samariya2023, DeMedeiros2023}. In addition, the challenge of transmitting large amounts of data to external servers for model training hinders practicality, making a shift to edge computing essential to keep bandwidth usage low and sensitive data close to the machine. Federated Learning (FL) is emerging as an alternative for use in resource-constrained industrial environments, by bringing operations closer to the source - the machine itself - to enable on-device training and inference, preserving privacy while ensuring real-time anomaly detection \cite{Abreha2022,Wang2023}. Unlike traditional centralized approaches, FL works on the principle of model training, where the data is not shared, but the models are. FL works in a distributed device scheme, using collective intelligence to train specific ML models aggregated in a central server while preserving the privacy of machine data \cite{Wang2023}. Thus, in addition to leveraging the on-device training paradigm and updates from distributed devices, FL also minimizes the communication overhead typically associated with transmitting large datasets to a central server.

Amidst the discussions on FL in anomaly detection for industrial scenarios, our study aims to contribute with answers to the following crucial aspects:
\begin{itemize}
    \item What are the requirements for implementing FL within an LP-WAN-based architecture, such as LoRaWAN, used for machine management? 
\end{itemize}
\begin{itemize}
    \item How can the limitations and constraints inherent to LoRaWAN, such as message limits and periodicity of data collection, be overcome to avoid compromising the accuracy of anomaly detection models using FL?
    
    \item What are the key strategies for maintaining the performance of FL with accuracy levels similar to those of centralized models?

\end{itemize}

Going deeper, optimising aggregation algorithms to minimise communication within the LoRaWAN framework is essential. In addition, addressing the data imbalance inherent in LoRaWAN schemes where some devices operate with different periodicity is critical to generalizing federated models in IIoT-based environments \cite{Aggarwal2023}. Finally, given that anomaly detection models predominantly rely on unsupervised approaches, leveraging human expertise to move towards semi-supervised techniques (by intervening in the federated training loop to provide feedback and refine models) holds promise for improving model performance in the future.

The contributions of this study focus on critical aspects of the use of FL for anomaly detection in machinery (in this case study used in the civil construction industry):
\begin{itemize}
    \item An anomaly detection framework based on autoencoder-federated learning is proposed. It uses monitoring data from heavy-duty machinery based on IIoT with LoRaWAN.
    \item A comparative analysis is conducted to evaluate the performance of federated models in anomaly detection against centralized models.
    \item The challenges of training models within a LoRaWAN-based scenario are analyzed, considering the training times when modifying the configuration of final rounds and parameter size. Hence, it is analyzed how the modifications of the model's structure affect the number of LoRaWAN messages to complete the training, thereby increasing the required time. 
\end{itemize}

The rest of the article proceeds as follows: section \ref{sec:RelatedWorks} presents the related work for our research, section \ref{sec:Background} overviews anomaly detection and its integration within FL processes. Section \ref{sec:DataCollection} details the process of collecting and preprocessing real industrial machinery data, focusing on approaches for anomaly labeling. Section \ref{sec:Centralized} delves into implementing centralized anomaly detection models, utilizing Isolation Forest (IF), One-Class Support Vector Machines (OC-SVM), and autoencoder (AE) as centralized models alongside their respective optimizations. Section \ref{sec:Federated} elaborates on developing a FL framework for anomaly detection. Section \ref{sec:experiment} presents the results from analyzing the framework and delves into the contextual implications within LoRaWAN architectures. Finally, the article concludes in section \ref{sec:Conclusion} by discussing future research directions and conclusions.

\section{Related work}
\label{sec:RelatedWorks}

An artificial neural network architecture named autoencoder has been a great ally in anomaly detection due to its ability to identify irregular patterns within different data sets. Today, significant progress has been made in the literature regarding FL schemes and autoencoders, particularly within the industrial domain. In addition, work is being done to find optimized models by making them smaller and with lower computational power requirements so they can be used in more constrained environments. Some work on anomaly detection using FL and autoencoders is presented below.

Becker et al. \cite{Becker2022} propose an unsupervised anomaly detection approach based on a reconstruction error model using autoencoders and FL to monitor the state of motors using IIoT. They perform an extensive evaluation compared to centralized models on two real-world datasets and several testbeds. Their results point to competitive performance and reduced resource and network utilization by up to 99.22\%, benefiting remote industrial sites. Truong \cite{Truong2022} proposes a robust distributed anomaly detection architecture for industrial control systems using hybrid FL, autoencoder, transformer, and Fourier mixing sublayer design. The FL framework designs a method to use edge sites to share model information for optimized anomaly detection performance, allowing lightweight, low CPU and memory usage, and low communication costs. The method enables deployment on edge devices, demonstrating training times of less than 7 minutes, and \SI{1.4}{Kib/s} in average uplink bandwidth usage, and
1.33 {1.33}{Kib/s} for downlink, considering the case of 16 hidden layers. Furthermore, Qin \cite{Qin2020} proposes a selective model aggregation to improve anomaly detection accuracy, considering an approach that addresses the problem of insufficient training data on a single-edge device. The work calculates the prediction errors to exclude unsatisfactory local models, and the method outperforms FedAvg, obtaining values of F1-score higher than 0.946 against 0.866 of the regular scenario. Kea et al. \cite{Kea2023} also propose a method that combines autoencoder and FL for anomaly detection in distributed power systems. Their approach achieves high prediction accuracy (over 0.86 F1-score), and similar performance compared with state-of-the-art models. It uses dynamic threshold choice using the data that overcomes the peaks in a certain scenario in the data distribution for each round. Finally, Mohammadi et al. \cite{Mohammadi2023} evaluate FL using LSTM-AE (Long Short-Term Memory Autoencoder)-based models, outperforming centralized anomaly detection models. They conclude that a dynamic threshold selection for anomaly detection is a better solution for centralized models. Otherwise, hybrid methods such as FL-AE-OCSVM or FL-AE-IF models are better for anomalies close to normal data, achieving 99\% F1-score on synthetic datasets.


Although significant progress has been made in anomaly detection using FL, several issues still need to be addressed in the literature, considering the challenges of resource-constrained scenarios. To the best of our knowledge, our research is the first work that aims to analyze anomaly detection using FL in a constrained environment, such as LoRaWAN. Thus, this paper aims to fill a gap in the literature by presenting a comprehensive approach, starting with a framework to implement anomaly detection combining FL and autoencoder. In addition, a performance comparison is made with well-known algorithms such as OC-SVM, IF, and centralized AE. The fundamental aspects needed to implement these models in LoRaWAN environments are discussed. In addition, a significant contribution to the analysis is made using data sets obtained from construction machinery working on a real construction site in Vilar Formoso, Portugal. 

\section{Background}
\label{sec:Background}

\subsection{Overview of anomaly detection approaches}

IIoT devices facilitate early anomaly detection by continuously monitoring industrial processes in real-time. These devices detect abnormal machine behavior, identifying events, failures, or potential security attacks. Two main approaches have emerged in IIoT anomaly detection: statistical methods and machine learning techniques \cite{Rodriguez2023}. Statistical methods, such as interquartile range (IQR), local outlier factor (LOF), and density-based methods, are used to classify anomalies by quantifying the deviation of anomalous data from normal data patterns. These statistical techniques can be complemented by machine learning algorithms, including(OC-SVM), IF, and K-nearest Neighbors (KNN), which leverage the power of supervised and unsupervised learning to improve anomaly detection accuracy. Alternatively, machine learning approaches using neural networks such as Autoencoder (AE), Recurrent Neural Networks (RNNs), Long Short-Term Memory (LSTM), and Convolutional Neural Networks (CNNs) offer sophisticated modeling capabilities that are particularly suited to complex and high-dimensional data \cite{Rodriguez2023,Sikder2023,Samariya2023}. By leveraging neural network architectures, these methods can effectively capture intricate patterns and relationships within the data, enhancing the performance of anomaly detection systems in IIoT environments.

\subsection{Autoencoders}

AE is a type of neural network that is widely used in the literature for anomaly detection due to its ability to reconstruct input data and capture complex patterns within the data distribution \cite{DeMedeiros2023}. Autoencoders consist of an encoder and decoder network, aiming to reconstruct input data accurately \cite{Ahn2023}. When presented with anomalies during testing, the model struggles to reconstruct such instances accurately due to their dissimilarity to normal data, resulting in higher reconstruction errors. AE models are particularly well-suited for resource-constrained environments because they efficiently compress and reconstruct data representations. Unlike complex neural network architectures that may require significant computational resources, AE models can be relatively lightweight and suitable for compression and operate efficiently on constrained devices with limited processing power and memory \cite{Abadade2023}.

\subsection{Federated Learning for anomaly detection}

FL is a collaborative training approach where devices participate in localized model training, using their respective datasets and sending local model weights to an aggregating central server. Model aggregation in FL consolidates information from multiple decentralized devices obtained during each FL round, thus contributing to a generalized global model with enhanced predictive capabilities\cite{Wang2023,Zhang2021}. Common aggregation methods include FedAvg, FedProx, FedNova, scaffold, Zeno, and Per-FedAvg. The selection of an aggregation algorithm in FL hinges on the specific needs of the application. Factors such as the complexity of aggregation, the necessity for tailored updates, adaptability in learning rates, device clustering, and the need for dynamic convergence influence the choice of algorithm \cite{Qi2024,An2023,Shi2022, Karimireddy2020,Wan2021}. One possibility is to use a neural network-based anomaly detection method that joins AE with the advantages of a decentralized framework like FL. In this context, IIoT devices can autonomously train well-defined models for anomaly detection, preserving privacy and minimizing the data transmission required by centralized models \cite{Khan2021,Chen2021}.

\subsection{Flower: A Framework for emulation Federated Learning clients}
This paper is based in part on the emulation of IIoT prototypes and FL implementation using Flower, an open source Python-based framework that is a key enabler in FL.  In essence, Flower provides a framework that orchestrates the collaborative training of machine learning models across distributed devices.  This framework operates through a client-server architecture, where clients perform local model training on their respective data, while periodically synchronising model updates with a central server.  One of the key advantages of Flower is its implementation flexibility, which allows the implementation of federated learning algorithms with different aggregation strategies and the ability to customise training protocols to suit specific use cases.  In addition, Flower supports several machine learning frameworks, allowing seamless integration with existing models and workflows, such as Keras for implementing autoencoders\cite{Beutel2022}.

\section{Data Collection and Preprocessing}
\label{sec:DataCollection}

\begin{figure}[t]
\centering
\includegraphics[width=2.3 in]{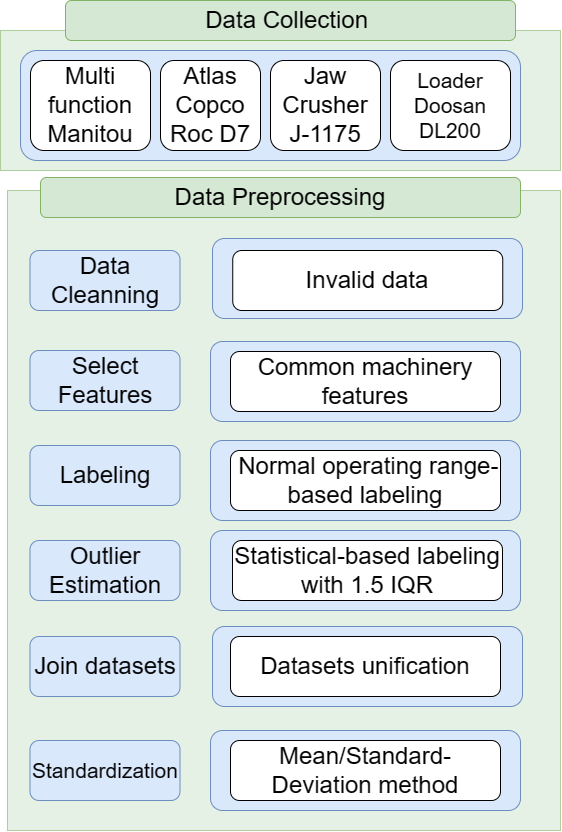}
\caption{Data Collection and Preprocessing}
\label{fig:DataCollection}
\end{figure}

As mentioned earlier, one of the objectives of this study was to conduct a comparative analysis between centralized and FL models. To provide context for this analysis, it is essential to outline the datasets utilized in this study, which will be employed in evaluating centralized and FL models. This section comprehensively describes the data collection and preprocessing procedures used throughout the study, depicted in Figure \ref{fig:DataCollection}.

\subsection{Data Collection}

\begin{figure}[t]
\centering
\includegraphics[width=3.2 in]{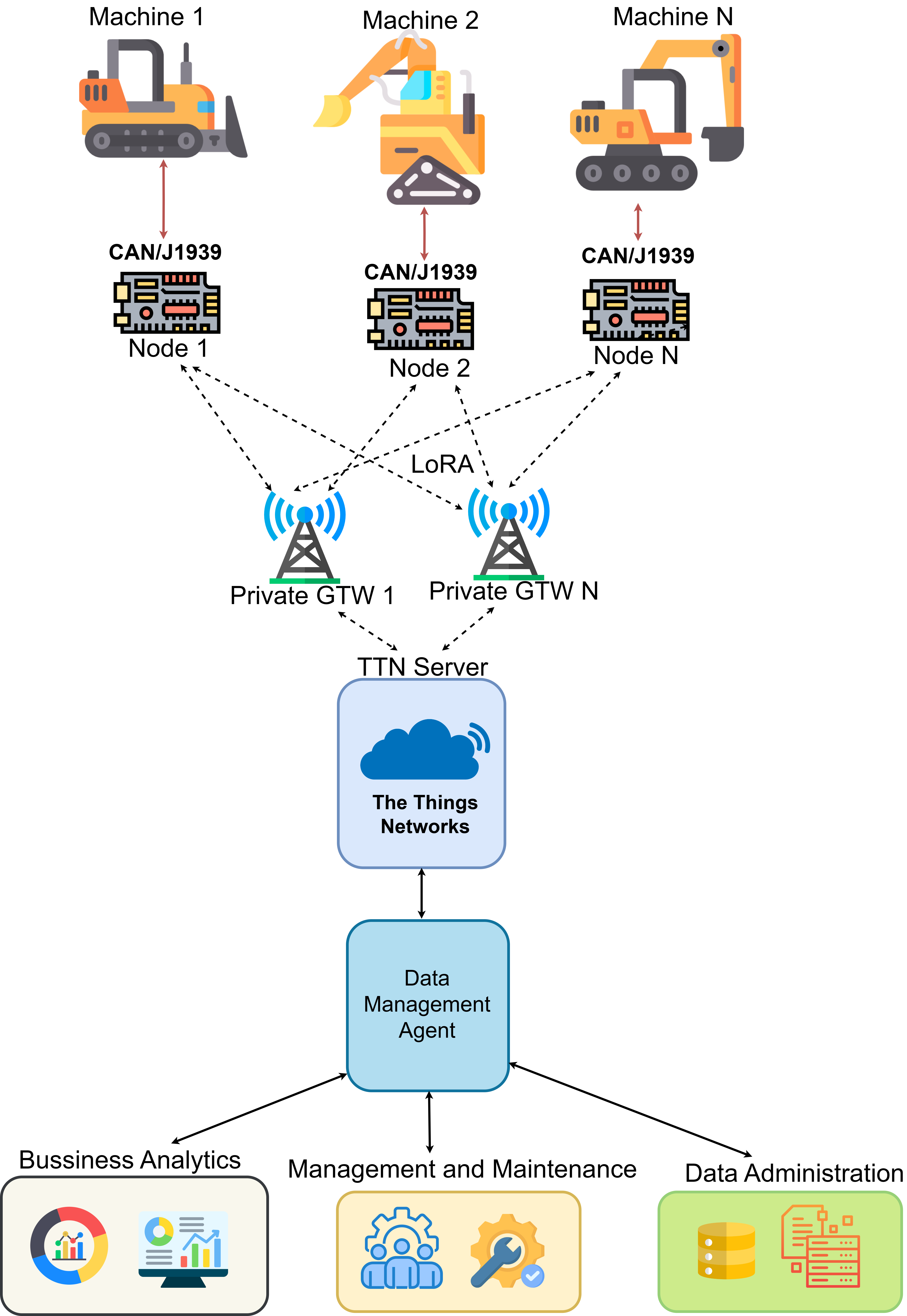}
\caption{IIoT-based management architecture}
\label{fig:architecture}
\end{figure}

The data collection procedure initially entails gathering construction industry machinery data through the approach described in \cite{Sanchez2023}. The data collection and monitoring framework, illustrated in Figure \ref{fig:architecture}, primarily relies on a private LoRaWAN network comprising four devices that monitor civil construction machinery via the CAN/J1939 protocol. Subsequently, a data management agent stores and manages the collected data, establishing a connection to The Things Network (TTN) server using MQTT.

\begin{figure}[t]
\centering
\includegraphics[width=2.3 in]{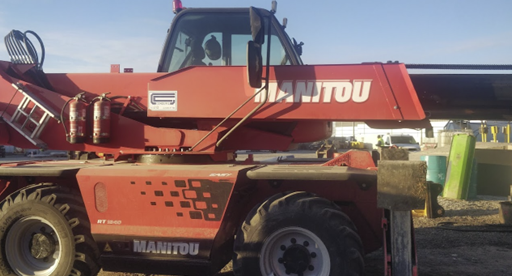}
\caption{Multifunctions Manitou}
\label{fig:Multifunctions}
\end{figure}

\begin{figure}[t]
\centering
\includegraphics[width=2.3 in]{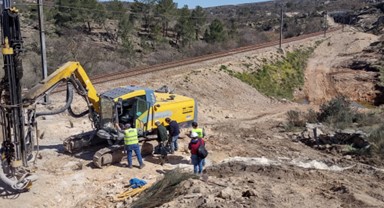}
\caption{Crawler drill Atlas Copco Roc D7}
\label{fig:AtlasD7}
\end{figure}

\begin{figure}[t]
\centering
\includegraphics[width=2.3 in]{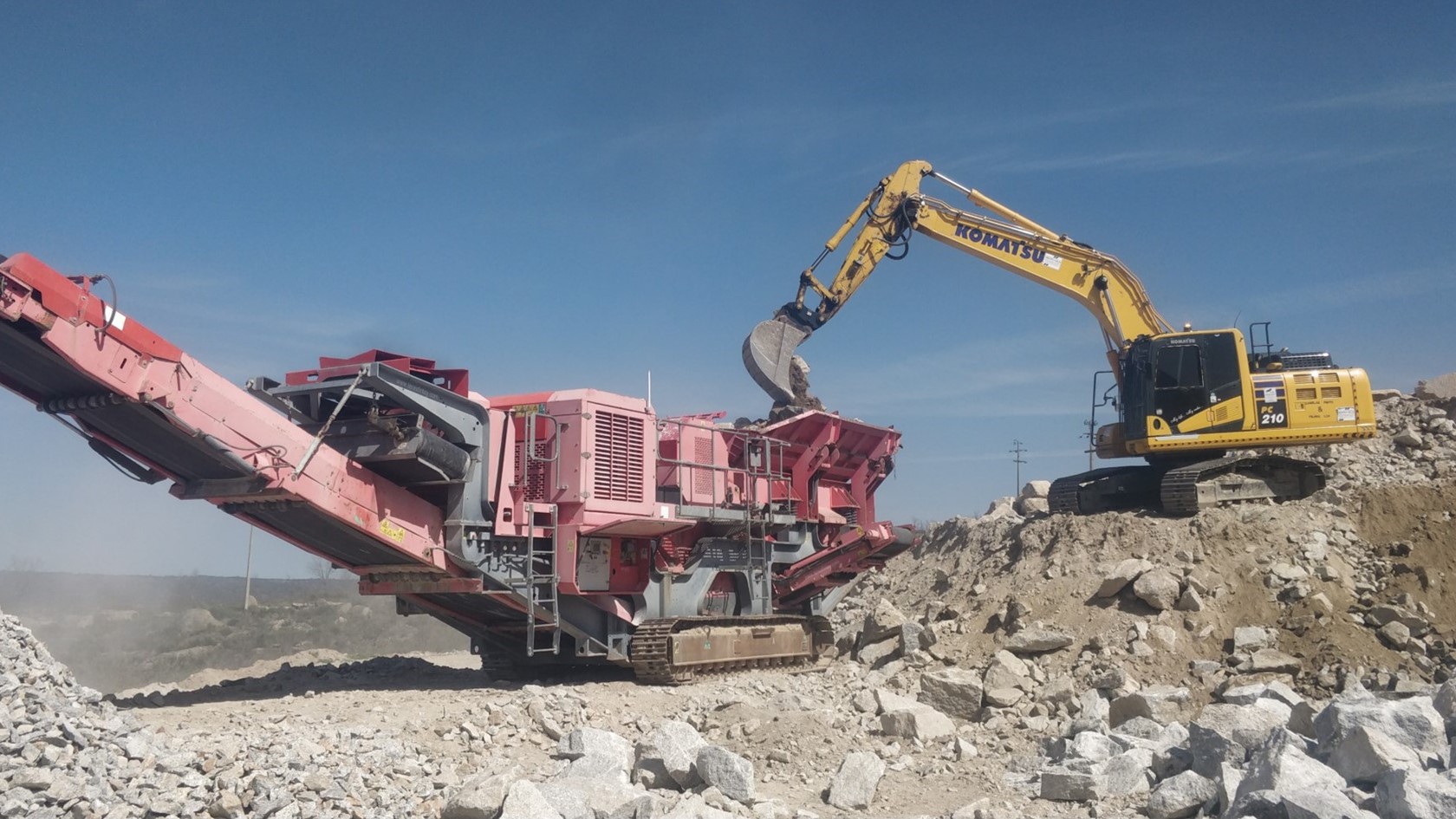}
\caption{Jaw Crusher terex J-1175}
\label{fig:JawCrusher}
\end{figure}

\begin{figure}[t]
\centering
\includegraphics[width=2.3 in]{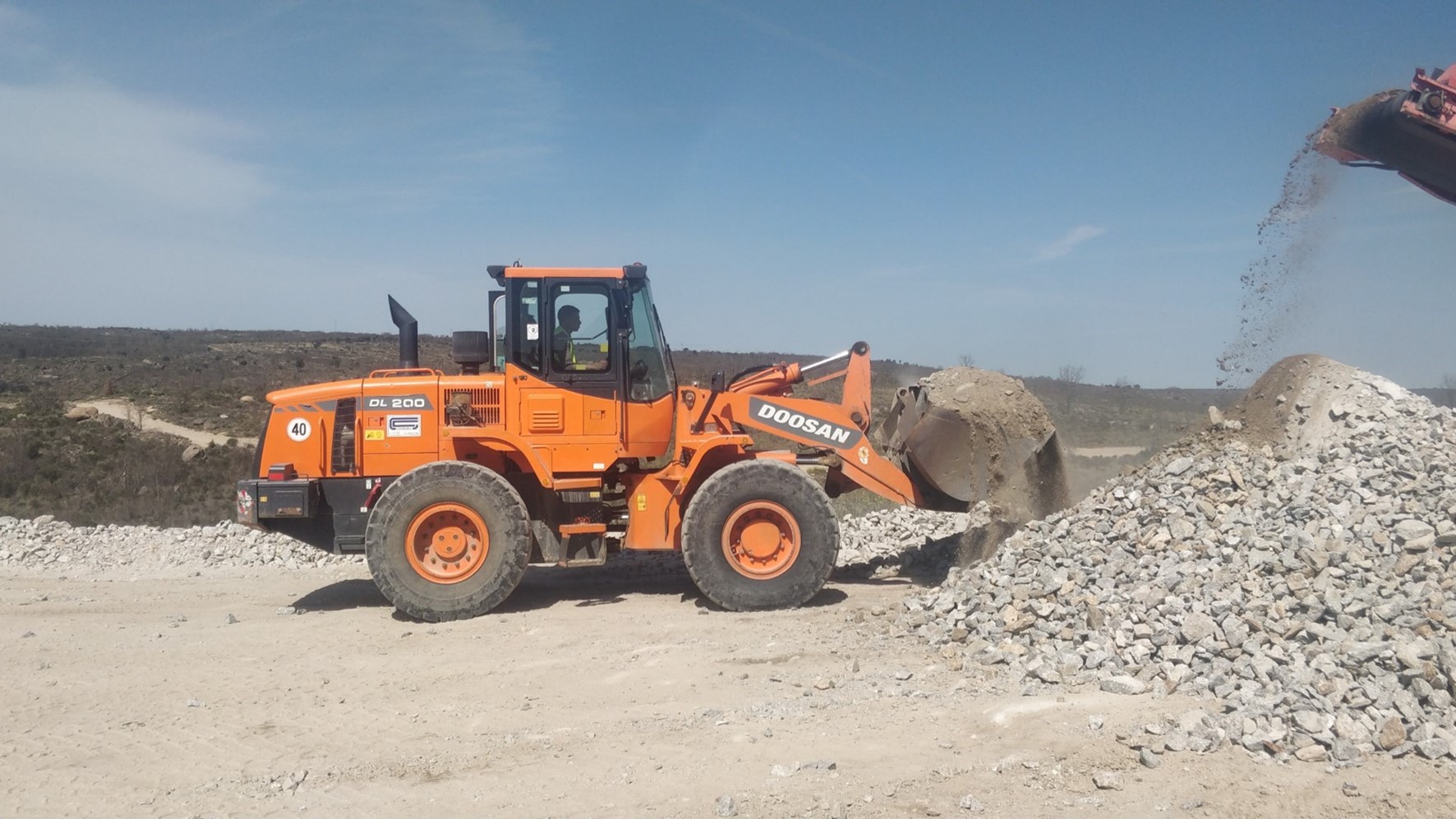}
\caption{Wheel loader Dossan DL200}
\label{fig:DoosanDL}
\end{figure}

The machines involved in the experiments and data collection are 
Multifunctions Manitou (Figure \ref{fig:Multifunctions}), Crawler drill Atlas Copco Roc D7 (Figure \ref{fig:AtlasD7}), Jaw Crusher terex J-1175 (Figure \ref{fig:JawCrusher}), and wheel loader Dossan DL200 (Figure \ref{fig:DoosanDL}). The experimental setup comprises two phases of prototype installation. The initial phase started on March 1, 2023, with the installation of the prototype on the Multifunctions machine, followed by the installation on Atlas Copco, Jaw Crusher, and Wheel loader Doosan. The number of messages received during the period of the experiment was 77798; the messages were standardized and contained readings of the J1939 protocol, such as battery voltage, fuel consumption, revolutions per minute (RPM), engine water and oil temperature, and oil pressure, in addition to the prototype’s readings, such as location, device identification, accelerometer information, and traveled distance.

\subsection{Data Preprocessing}
\label{sec:DataPreprocessing}


After monitoring the machines and extracting performance data during normal operation, the next crucial step was to preprocess this data for integration into the machine learning workflow. The following subsections outline the specific preprocessing procedures customized for the data and individual machines. This essential phase refines and structures the data, establishing a solid groundwork for subsequent machine-learning tasks.

\subsubsection{Data Cleaning}

The purpose of data cleaning is to ensure correct data for each machine, where only valid data centered on the controlled periods of experimentation remain. For example, the presence of “invalid data” in the J1939 protocol, represented by ”FF” values,” should not be taken into account in the process, nor should the invalid configuration epoch. After cleaning, the Multifunctions Manitou machine had 10150 data instances, the Doosan DL200 machine had 11507 data instances, the Jaw Crusher machine had 6677 data instances, and the Atlas Copco machine had only 388 data instances.

\subsubsection{Features selection}

Once the data was cleaned, they were analyzed exhaustively, obtaining the J1939 features that were decisive to the operation of the machine during the considered timeframe. The selected features were those in which the four machines presented valid values, namely, potential, RPM, consumption, engine water temperature, and oil. These represent critical parameters consistently present among the various machines. Table \ref{tab:features} summarizes the obtained J1939 features. This focused selection streamlines the analysis process and ensures a standardized approach to anomaly identification, leveraging shared critical features to understand the operational status and potential irregularities across multiple machines.

\subsubsection{Normal operating range-based labeling}

The labeling process for establishing normal range values involved a meticulous analysis of each machine’s characteristics, incorporating insights from the manufacturers’ manuals and authoritative blogs detailing typical operational behaviors. Delving into detailed technical documentation and leveraging insights from expert knowledge sources provided a robust foundation for defining the normal behavior of the selected features. Table \ref{tab:normalvalues} summarizes the considered normal values used for labeling the dataset for each machine; labels are used as the basis for a semi-supervised approach to improving centralized and federated models. Considering this, an aggregating labeling process was used to obtain a single instance label. The method used classified an instance as anomalous if at least one of its features is flagged as anomalous, aiming to consolidate the individual anomaly labels from multiple features into a unified label for each instance.

\begin{table}[!t]
\setlength\extrarowheight{2pt}
\centering
\resizebox{0.72\columnwidth}{!}{\begin{tabular}{l|l|l|l|l} 
\hline
\begin{tabular}[c]{@{}l@{}}\textbf{Machine / }\\\textbf{Signals}\end{tabular} & \textbf{Manitou}                             & \textbf{D7 }                                 & \textbf{J-1175 }                             & \textbf{DL200 }                               \\ 
\hline
\textbf{Battery}                                                              & \begin{tabular}[c]{@{}l@{}}Ok\\\end{tabular} & \begin{tabular}[c]{@{}l@{}}Ok\\\end{tabular} & \begin{tabular}[c]{@{}l@{}}Ok\\\end{tabular} & \begin{tabular}[c]{@{}l@{}}Ok\\\end{tabular}  \\ 
Consumption                                                                        & \begin{tabular}[c]{@{}l@{}}Ok\\\end{tabular} & \begin{tabular}[c]{@{}l@{}}Ok\\\end{tabular} & \begin{tabular}[c]{@{}l@{}}Ok\\\end{tabular} & \begin{tabular}[c]{@{}l@{}}Ok\\\end{tabular}  \\ 
Hours                                                                         & \begin{tabular}[c]{@{}l@{}}Ok\\\end{tabular} & \begin{tabular}[c]{@{}l@{}}NA\\\end{tabular} & \begin{tabular}[c]{@{}l@{}}Ok\\\end{tabular} & \begin{tabular}[c]{@{}l@{}}NA\\\end{tabular}  \\ 
RPM                                                                      & \begin{tabular}[c]{@{}l@{}}Ok\\\end{tabular} & \begin{tabular}[c]{@{}l@{}}Ok\\\end{tabular} & \begin{tabular}[c]{@{}l@{}}Ok\\\end{tabular} & \begin{tabular}[c]{@{}l@{}}Ok\\\end{tabular}  \\ 
\textbf{Oil (ºC)}                                                             & \begin{tabular}[c]{@{}l@{}}NA\\\end{tabular} & \begin{tabular}[c]{@{}l@{}}NA\\\end{tabular} & \begin{tabular}[c]{@{}l@{}}NA\\\end{tabular} & \begin{tabular}[c]{@{}l@{}}NA\\\end{tabular}  \\ 
\textbf{Water~\textbf{(ºC)}}                                                  & \begin{tabular}[c]{@{}l@{}}Ok\\\end{tabular} & \begin{tabular}[c]{@{}l@{}}Ok\\\end{tabular} & \begin{tabular}[c]{@{}l@{}}Ok\\\end{tabular} & \begin{tabular}[c]{@{}l@{}}Ok\\\end{tabular}  \\ 
\textbf{Oil (Bar)}                                                            & \begin{tabular}[c]{@{}l@{}}Ok\\\end{tabular} & \begin{tabular}[c]{@{}l@{}}Ok\\\end{tabular} & \begin{tabular}[c]{@{}l@{}}Ok\\\end{tabular} & \begin{tabular}[c]{@{}l@{}}Ok\\\end{tabular}  \\ 
Fuel                                                                          & \begin{tabular}[c]{@{}l@{}}NA\\\end{tabular} & \begin{tabular}[c]{@{}l@{}}NA\\\end{tabular} & \begin{tabular}[c]{@{}l@{}}NA\\\end{tabular} & \begin{tabular}[c]{@{}l@{}}NA\\\end{tabular}  \\
\hline
\end{tabular}}
\caption{Valid Features for each machine}
\label{tab:features}
\end{table}

\begin{table}[t]
\setlength\extrarowheight{2pt}
\centering
\resizebox{0.95\columnwidth}{!}{\begin{tabular}{l|l|l|l|l} 
\hline
\begin{tabular}[c]{@{}l@{}}\textbf{Machine / }\\\textbf{Signals}\end{tabular} & \textbf{Manitou } & \textbf{D7 } & \textbf{J-1175 } & \textbf{DL200 }  \\ 
\hline
Battery [Voltage]                                                                   & 12.6-13.6~      & 24-28      & 24-28          & 24-28          \\ 
Consumption [l/h]                                                                  & 1-40~           & 1-40~       & 1-40~          & 1-40~          \\ 
RPM                                                                & 800-2200        & 800-2200   & 800-2200       & 800-2200       \\ 
Water [°C]                                                                    & 75-100          & 75-100     & 75-100         & 75-100         \\ 
Oil [Bar]                                                                     & 1-7             & 1-7        & 1-7            & 1-7            \\
\hline
\end{tabular}
}
\caption{Normal range values for labelling process}
\label{tab:normalvalues}
\end{table}

\subsubsection{Statistical-based labeling}

This framework combines ML models with a statistical approach, such as the 1.5 IQR method for anomaly detection, ensuring complete coverage of outlier detection. By integrating both approaches, we can leverage the strengths of each method: the robustness of statistical techniques such as the 1.5 IQR method and the predictive power of ML models. Data from each machine are used to calculate the lower and upper bound as a function of interquartile distance, and label data that exceed the thresholds. After labeling and considering all machines, the data set was found to have 16.44\% of data outliers. The outlier labeling was used for the initial estimation of the hyperparameters of the ML models.

\subsubsection{Join Datasets}

In this framework, consolidating machine data into a single dataset serves multiple purposes. This unified dataset enables comprehensive analysis and improves centralized models' performance, encourages generalization, and seeks greater accuracy and scalability across machines. In addition, it allows using features from all machines to generate training, validation, and test data sets.

\subsubsection{Machine Value Standardisation}

The combined dataset was standardized using the five selected features' with mean ($\bar{x}$) standard deviation ($\sigma$) method. Standardization consisted of scaling the features to have a $\bar{x}$ of zero and a $\sigma$ of one, making the data fit a standard, normal distribution. This process preserves the shape of the original distribution, maintaining the relative relationships between data points. This stage is especially beneficial in anomaly detection tasks, where accurately identifying deviations in readings from multiple sensors on multiple machines is vital for effective model training and accurate anomaly prediction.

\begin{figure}[!t]
\centering
\includegraphics[width=2.3 in]{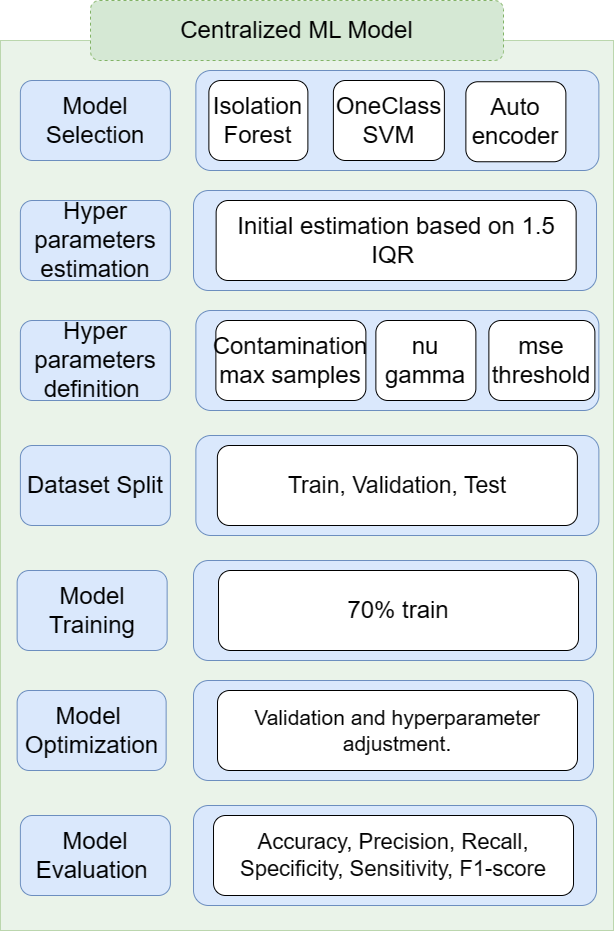}
\caption{Centralized models pipeline}
\label{fig:Centmodels}
\end{figure}

\section{Centralized Anomaly Detection Models Implementation and Analysis}
\label{sec:Centralized}

After preprocessing the data, the next step was to use them in anomaly detection models. Optimizing the centralized models, which serve as a benchmark for evaluating FL performance, is crucial before incorporating them into FL frameworks. Figure \ref{fig:Centmodels} shows the stages of centralized model implementation, which are described in the following subsections.

\subsection{Model Selection and hyperparameters}

The comparison with centralized models encompassed two non-neural-network-based algorithms, IF and OC-SVM, alongside a neural-network-based model, AE. IF employs a random feature selection and a recursive partitioning approach, isolating data points until they are individually segmented. At the same time, OneClassSVM delineates a decision boundary around normal data points in the feature space. Both methods are unsupervised anomaly detection methods. The 16.44\% of anomalies identified by the IQR method are initially utilized for parameter initialization on IF and OC-SVM. For IF, parameters such as contamination, representing the expected anomaly proportion in the dataset, and max samples, determining the data points used in constructing each tree, are used. Conversely, OC-SVM parameters include $\mu$, denoting the center of the hypersphere used for normal-anomaly separation, and $\gamma$, reflecting the influence of individual training samples on the decision boundary.

\begin{figure}[t]
\centering
\includegraphics[width=3.2 in]{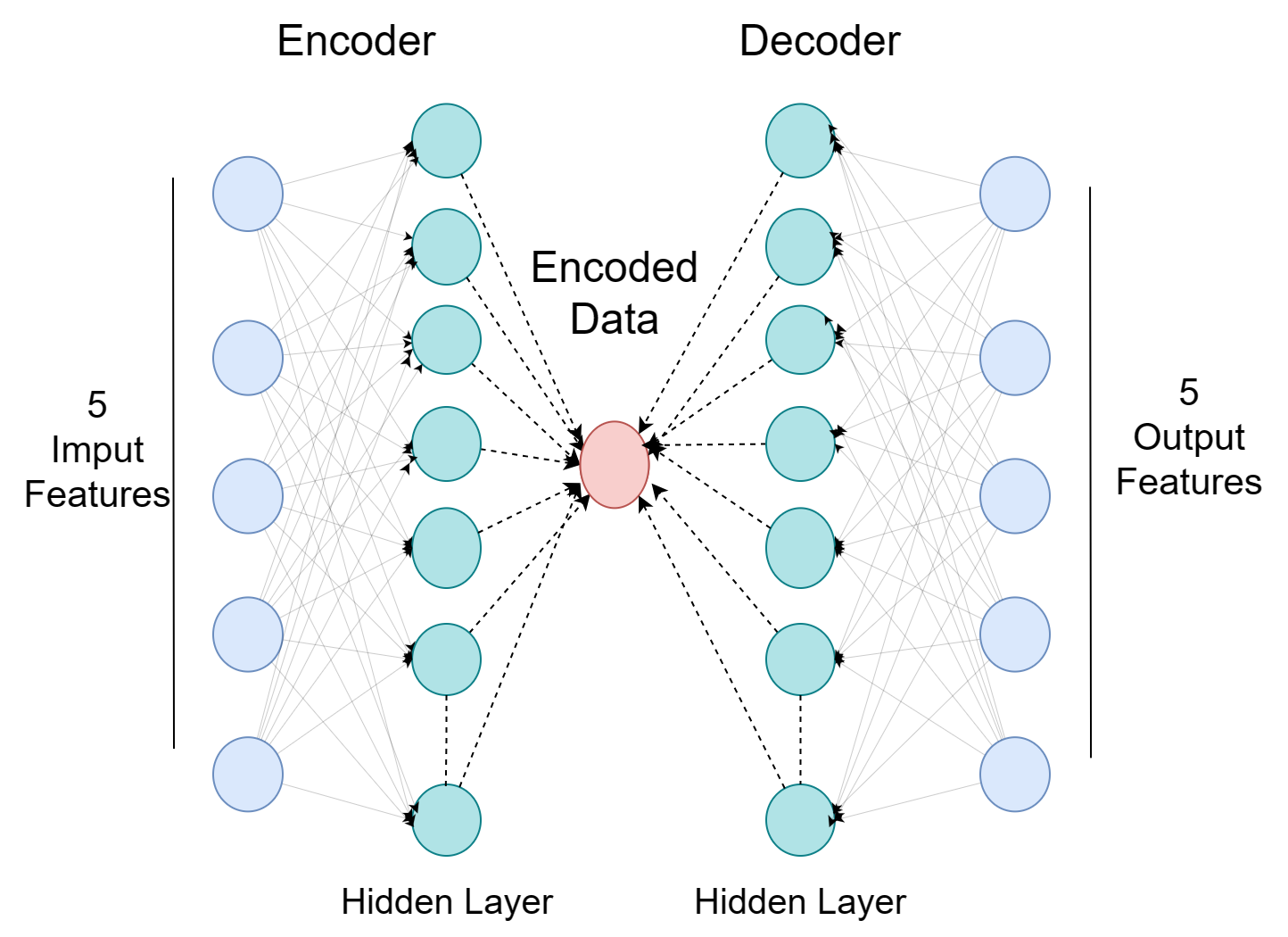}
\caption{Autoencoder structure}
\label{fig:autoencoderStruct}
\end{figure}

Unlike non-neural-network-based algorithms, which rely on predefined hyperparameters derived from 1.5 IQR, an autoencoder works differently in anomaly detection by reconstructing the data within the neural network. The autoencoder has a neural network architecture similar to a ”bottleneck” structure consisting of an encoder/decoder, as shown in Figure \ref{fig:autoencoderStruct}. For these, the input consists of 5 neurons representing the selected features, followed by a hidden layer with initially 32 neurons, and a ReLU activation function, which promotes sparse representations by reducing negative values to zero. The structure is maintained by inverting it for the decoder, where the input will be the output of the encoder, having five output neurons. This structure is kept as simple as possible, i.e., with a single hidden layer to keep the size of the parameters as small as possible so that they can be used in LoRaWAN scenarios. However, the neural network structure parameters may be optimized in future steps.
Furthermore, the model is compiled using the Adaptive Moment Estimation (Adam) optimizer, and Mean Squared Error (MSE) is used as a loss function for model improvement. A crucial aspect of anomaly detection with autoencoders is the selection of the evaluation threshold to label or not a value as an anomaly. In this instance, MSE is computed as the average squared deviation between the actual and predicted values, as in equation \ref{eq:mse}. In this initial approximation, the squared deviation of each column is utilized, and the value representing the 84th percentile boundary of these deviations serves as the threshold. By extracting the 84th percentile of this deviation, we acknowledge the presence of 16\% outliers observed within the 1.5 IQR method.
\begin{equation}
\text{MSE} = \frac{1}{n} \sum_{i=1}^{n} (y_i - \hat{y}_i)^2
\label{eq:mse}
\end{equation}

\subsection{Dataset Split}

\begin{figure}[t]
\centering
\includegraphics[width=3.2 in]{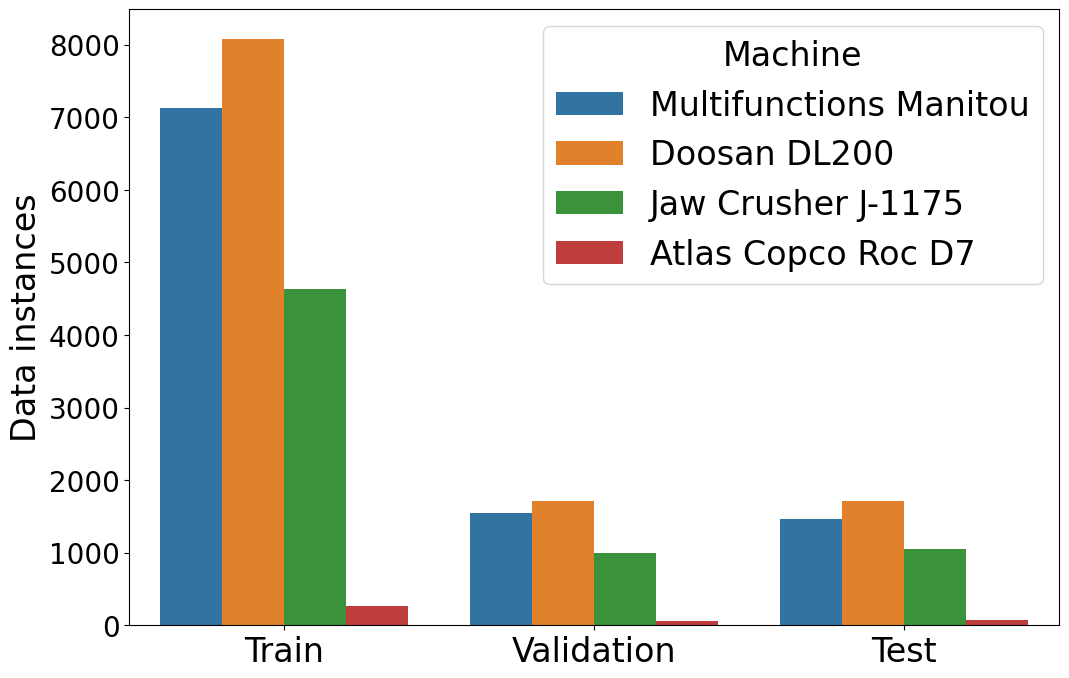}
\caption{Data distribution after split}
\label{fig:dataDist_split}
\end{figure}

Once the initial parameters have been estimated, training these models to detect anomalies and evaluate their performance is necessary. This study uses 70\% of the data for training, 15\% for validation and future optimization, and 15\% for testing. Sklearn Tools split the data randomly but kept the proportion between machines. Therefore, the data distribution per machine and dataset is shown in Figure \ref{fig:dataDist_split}. 

\subsection{Model Optimization}
\label{sec:ModelOptimization}
After defining the data sets, the next crucial step is models evaluation and optimization. Optimization is achieved by grid-search, a systematic method that iteratively explores combinations of hyperparameters to find the optimal parameters of the centralized models for both non-neural and neural network-based. In the case of IF, parameters such as contamination and max samples are considered. For contamination, arrays of values ranging from 1\% of expected anomalies to 20\% are examined, while for max\_samples, which are the percentage of data used to create the trees, arrays from 10\% to 50\% of the samples are used. Now, for OC-SVM, the $\nu$ and $\gamma$ parameters are tuned; the $\nu$ parameter is similar to the contamination in an IF; we use a percentage between 1-20 \% of expected anomalies and an array in logarithmic scale from 0.001 to 0.1 in $\gamma$, which allows to cover larger ranges in the decision boundaries.

\begin{figure}[t]
\centering
\includegraphics[width=2.6 in]{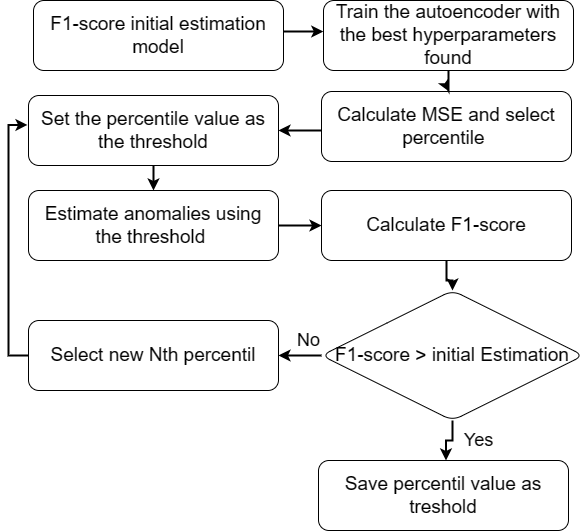}
\caption{Threshold selection process}
\label{fig:treshold}
\end{figure}

In addition, the neural network’s optimization and the threshold selection for the autoencoder are performed. For better optimization of the autoencoder, the model is defined with its score metric, corresponding to the negative value of the MSE calculated between the predicted data and the original data, taking into account the validation dataset for the optimization. Using this proprietary metric outside the conventional metrics allows us to optimize parameters that enable us to reconstruct the data better since they will highlight those values considered anomalies. Now, for the parameters within the grid, we took into account values of hidden layer size between 16 and 128, considering the same size for encoding and decoding, epochs between 10 and 100, batch size between 16 and 128, and different activation functions such as ”ReLu,” ”Tanh,” and ”Sigmoid.” Finally, selecting the appropriate threshold for anomaly detection is crucial to optimize the anomaly detection process. Figure \ref{fig:treshold} illustrates the process of determining the optimal threshold value using the validation datasets. For this process, initially, the F1-score was calculated for the validation dataset based on the training dataset and initial hyperparameters, serving as a reference for threshold selection. Subsequently, predictions were generated for the validation dataset using the best-performing model identified through grid search. The MSE is then computed as the mean of the squared differences between the actual test data and the predictions for each value. Next, a range of threshold values was iterated, derived from previously calculated percentile values of the MSE. These threshold values were used to label anomalies, enabling the calculation of the F1 score. The aim was to maximize the F1-score, always comparing it against the F1-score obtained with the initial parameter estimate. The threshold yielding the highest F1 score was chosen as the optimal threshold for our process. This selected threshold value served as the ideal threshold for subsequent evaluations. Table \ref{tab:optimization} summarizes the grid search values and the best parameters discovered during optimization.

\begin{table}
\setlength\extrarowheight{2pt}
\centering
\begin{tabular}{l|l|l} 
\hline
\textbf{Model}       & \textbf{Grid Search Parameters}                                             & \textbf{Best}~  \\ 
\hline
IF                   & Contamination: [0.01-0.2]                                                   & 0.07            \\ 
                     & Max Sample: [0.1 - 0.5]                                                   & 0.27            \\ 
\hline
OC-SVM               & $\mu$ :[0.01-0.2]                                                             & 0.1167          \\ 
                     & $\gamma$: [0.001 - 0.1]                                                       & 0.01            \\ 
\hline
\textbf{Autoencoder} & Hidden layer size: [16-128]                                                & 32              \\ 
                     & Epoch: [10-100]                                                            & 80              \\ 
                     & Batch size: [16-128]~                                                      & 16              \\ 
                     & \begin{tabular}[c]{@{}l@{}}Activation: [ReLu,Tanh,Sigmoid] \end{tabular} & Tanh            \\ 
                     & Threshold value                                                             & 0.1622          \\
\hline
\end{tabular}
\caption{Optimization parameter and best values found.}
\label{tab:optimization}
\end{table}

\section{Federated Learning for anomaly detection}
\label{sec:Federated}
\begin{figure}[!t]
\centering
\includegraphics[width=2.2 in]{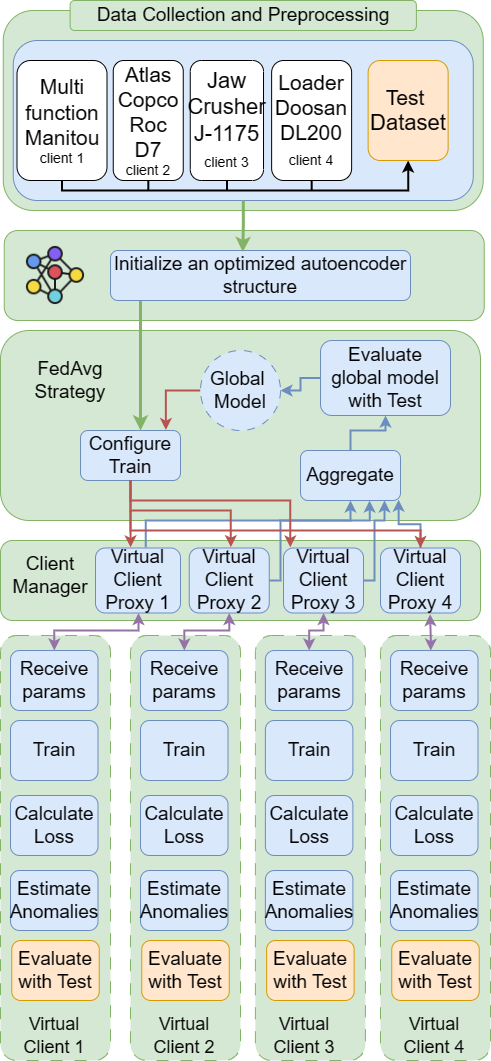}
\caption{Federated Learning Framework}
\label{fig:federatedframework}
\end{figure}

This section explores FL as a potent framework for anomaly detection in industrial settings. The section \label{sec:Centralized} shows the optimization of centralized models by having optimized models serve first as benchmarks for the subsequent evaluation of FL methodologies and, secondly, to have an optimized autoencoder structure for use in FL. Subsequently, we delve into the intricacies of implementing FL and enhancing model performance to deal with the requirements of client devices. The details of this implementation will be comprehensively presented in the forthcoming subsections based on the pipeline shown in Figure \ref{fig:federatedframework}.

\subsection{Client Data Preprocessing}

As Section \ref{sec:DataPreprocessing} outlines, the data undergo preprocessing before integration into anomaly detection models. Notably, each machine retains its distinct dataset in the FL process, diverging from the conventional approach, where data is amalgamated into a single training dataset. Nevertheless, a test dataset is preserved for comparative analysis with centralized models.

\subsubsection{Autoencoder initialization}

The FL process must use a known neural network structure that will be sent to each client to start the collaborative training. For this process, we use the parameters and structure of the optimized autoencoder described in section \ref{sec:ModelOptimization}. This approach lays a solid base for model convergence and performance across decentralized devices. It can mitigate issues such as overfitting or underfitting, which are common challenges in FL due to heterogeneous data distributions across clients. Starting with an optimized structure, as shown in Table \ref{tab:optimization}, allows the FL process to proceed more smoothly, leading to faster convergence and improved model accuracy.

\subsubsection{Virtual client initialization and orchestration}

FL virtual clients start the optimized autoencoder with random weights, allowing them to begin with an unbiased model and ensuring fairness across devices. The virtual client proxy executes the process, which acts as a representative entity of a simulated client participating in the initial configuration. The virtual client contains the client manager, which acts as the intermediary responsible for orchestrating communication between the simulated client and the central server during FL iterations. Second, through the iterative aggregation process, in which updates from individual clients are combined to form a global model, the random initialization of clients encourages harmonizing the model. As clients train their models locally on specific data sets, the subsequent aggregation process allows the central server to aggregate the updates.

\subsubsection{Aggregation Strategy}

The process uses a FedAvg strategy to aggregate model updates from multiple clients by averaging their respective model parameters. This allowed us to maintain a robust model even when unbalanced data sets exist from different machines. When clients have varying amounts or distributions of normal and abnormal data, the FedAvg approach can maintain substantial model performance. Aggregation of the models from each machine is performed within the server and then passed to the evaluation, and the model phases are updated for the clients. The process begins with initializing a global model $\theta_{0}$, distributed to all participating clients. Each client independently trains the model on its local data, generating model updates $\Delta \theta_i$. These updates are then aggregated at the server using a weighted average, where the weights correspond to the number of samples on each client. The updated global model $\theta_{t+1}$ is obtained by combining these model updates. This process iterates over multiple rounds $N$ until convergence or a predefined stopping criterion is met, which, in our scenario, corresponds to the 80 epochs found in optimization of the centralized models described on section\ref{sec:ModelOptimization} and table \ref{tab:optimization}. The following equation describes the process:
\begin{equation}
 \theta_{t+1} = \sum_{i=1}^{N} \frac{n_i}{N} \cdot \Delta \theta_i
\end{equation}

\subsubsection{Global model transmission}

The global $\Delta \theta_i$ model is stored on the server and prepared to be sent to the clients, prepared for transmission to individual clients through model serialization. Model serialization involves converting the model into a format that can be efficiently transmitted over the network and reconstructed at the client's end. This serialized form of the model, which includes the model architecture and trained parameters, can be sent over the network to the client devices for further training. The model is then deserialized at the client's end, allowing it to continue learning based on the aggregated updates from all participating devices. 

\subsubsection{Model enhancement}

The aggregation of models takes place at the server level, with model tuning occurring primarily at the client level. Each client fine-tunes its model during local training, using the MSE as a loss function. Before aggregation at the centralized server, clients adjust their parameters based on local data and training, contributing refined models to the collaborative learning process.

\subsubsection{Anomaly estimation}

During the optimization phase of the centralized model using the autoencoder, determining the threshold for labeling anomalies or non-anomalies was a crucial step. Similarly, the same principle guides the labeling of values and estimating anomalies within the FL framework. Upon completion of the defined rounds, each client fine-tunes the anomaly estimation to identify the  MSE that yields the highest F1 score. This process mirrors the approach taken in the centralized model, leveraging an initial reference value of 0.16225 to refine anomaly estimation and achieve optimal performance.

\section{Experiment details and evaluation results}
\label{sec:experiment}

The initial evaluation phase involves a comprehensive comparison between the performance of centralized models and the FL framework implemented in this study. This comparison serves as a critical benchmarking exercise, shedding light on the efficacy and potential advantages of employing FL for anomaly detection in industrial settings. We aim to discern the strengths and limitations of FL in contrast to traditional centralized models. The FL framework in this experiment simulates distributed devices functioning as clients orchestrated by the Flower framework, with a central server responsible for aggregating the individual models. The computational environment hosting this simulation boasts an 11th Gen Intel(R) Core(TM) i7-1165G7 processor clocked at \si{2.80}{GHz}, equipped with \si{16}{GB}  of RAM. The experimental setup operates within a simulated Conda environment utilizing the Windows Subsystem for Linux (WSL). This setup emulates a distributed system where multiple devices, represented by virtual clients, collaboratively train machine learning models under the FL paradigm. The server orchestrates aggregating these locally trained models, fostering collaborative learning while considering the computational resources available within this simulated environment.

\subsection{Evaluation Metrics}

In evaluating the performance of our models in the context of anomaly detection, the convention is to label the ”positive” class as normal values (i.e., instances considered regular or non-anomalous) and the ”negative” class as anomalies (i.e., instances considered irregular or anomalous). This convention is also used in statistical contexts when talking about outliers. Therefore, in the confusion matrix: True Positive (TP): Instances correctly classified as normal (correctly identified normal values); False Positive (FP): Instances incorrectly classified as normal (missed anomalies); True Negative (TN): Instances correctly classified as anomalies (correctly identified anomalies); False Negative (FN): Instances incorrectly classified as anomalies (false alarms). To evaluate the performance of the models, a set of metrics has been employed, encompassing Accuracy (Acc), Precision (Pre), Specificity or True negative rate (TNR), Sensitivity or True Positive Rate (TPR) or recall and F1-score (F1). This multifaceted evaluation framework thoroughly examines each model's ability to accurately classify normal and anomalous instances, providing a nuanced understanding of their overall efficacy in anomaly detection across diverse industrial machinery data. Based on the parameter of the confusion matrix, the metrics are calculated as follows: 

\begin{equation}
\text{Accuracy (Acc)} = \frac{\text{TP + TN}}{\text{TP+FP+TN+FN}}  \times  100 
\end{equation}

\begin{equation}
\text{Precision (Pre)} = \frac{\text{TP}}{\text{TP + FP}} \times  100 
\end{equation}


\begin{equation}
\text{True Negative Rate (TNR)} = \frac{\text{TN}}{\text{TN+FP}} \times  100
\end{equation}

\begin{equation}
\text{True Positive Rate (TPR)} = \frac{\text{TP}}{\text{TP+FN}} \times  100
\end{equation}

\begin{equation}
\text{F1-score (F1)} = \frac{\text{TP}}{\text{TP + 1/2 (FP+FN)}} \times  100 
\end{equation}

\subsection{Federated Learning vs Centralized Models}

After defining the evaluation metrics, our focus shifts to comparing the performance of centralized models against that achieved using the FL framework. For the non-neural-network-based centralized models, multiple iterations (13) were conducted with varying random state seeds to ensure the reproducibility of results across different runs of the code. In the case of the autoencoder-based model, the random seed was also initialized before the model creation to ensure consistency across runs. In contrast, the FL process involved random initialization of clients and weights, a step within the Flower simulation framework. Once the desired number of iterations was set, evaluation data was extracted from the resulting confusion matrix for analysis and comparison.

\begin{table}[t]
\setlength\extrarowheight{2pt}
\centering
\begin{tabular}{l|l|l|l|l|l} 
\hline
Metric & Value  & OC-SVM         & IF             & AE             & AEFL            \\ 
\hline
Accuracy    & min    & 91.97          & 92.88          & 93.11          & 89.42           \\
       & max    & 92.16          & 94.17          & 93.15          & 93.15           \\
       & mean   & \textbf{92.06} & \textbf{93.46} & \textbf{93.13} & \textbf{92.30}  \\
       & std    & 0.06           & 0.42           & 0.02           & 1.16            \\
       & median & 92.06          & 93.46          & 93.11          & 92.88           \\ 
\hline
Precision    & min    & 93.02          & 93.10          & 92.98          & 92.51           \\
       & max    & 93.15          & 94.08          & 93.01          & 97.88           \\
       & mean   & \textbf{93.11} & \textbf{93.54} & \textbf{92.99} & \textbf{96.70}  \\
       & std    & 0.05           & 0.33           & 0.02           & 1.80            \\
       & median & 93.12          & 93.48          & 92.98          & 97.69           \\ 
\hline
TNR    & min    & 81.20          & 81.20          & 80.79          & 80.70           \\
  & max    & 81.62          & 83.96          & 80.87          & 93.62           \\
       & mean   & \textbf{81.49} & \textbf{82.44} & \textbf{80.82} & \textbf{90.65}  \\
       & std    & 0.15           & 0.91           & 0.04           & 4.30            \\
       & median & 81.54          & 82.29          & 80.79          & 93.04           \\ 
\hline
TPR    & min    & 95.95          & 97.17          & 97.85          & 92.64           \\
  & max    & 96.37          & 98.10          & 97.88          & 93.50           \\
       & mean   & \textbf{96.13} & \textbf{97.70} & \textbf{97.86} & \textbf{92.93}  \\
       & std    & 0.14           & 0.27           & 0.02           & 0.25            \\
       & median & 96.11          & 97.75          & 97.85          & 92.85           \\ 
\hline
F1-score     & min    & 94.52          & 95.18          & 95.38          & 92.66           \\
       & max    & 94.67          & 96.05          & 95.38          & 95.38           \\
       & mean   & \textbf{94.59} & \textbf{95.57} & \textbf{95.36} & \textbf{94.77}  \\
       & std    & 0.05           & 0.29           & 0.02           & 0.85            \\
       & median & 94.59          & 95.57          & 95.35          & 95.19           \\
\hline
\end{tabular}

\caption{Comparison between models}
\label{tab:comparisonmodels}
\end{table}

Table \ref{tab:comparisonmodels} provides an overview of the results obtained from our experimentation. The table distinguishes between the non-null neural network-based models, OC-SVM and IF, and the centralized Autoencoder-based model, labeled AE, alongside its FL counterpart, referred to as AEFL. This classification facilitates a clear comparison of the performance metrics across different anomaly detection approaches. The analysis reveals notable trends in the performance metrics across various anomaly detection models. The IF model stands out with the highest mean accuracy (93.46\%) and F1-Score of 96.05\%, with an average specificity of 82.44\%. This suggests robust detection capabilities, particularly in identifying true anomalies relative to normal instances.
Conversely, the OC-SVM model demonstrates remarkable consistency, exhibiting low $\sigma$ values, not higher than 0.15\% across multiple runs. Comparing neural-network-based autoencoders with non-neural-network models, similar performance trends emerge, with an average F1-score of 95.36\% and accuracy of 93.13\%, respectively. Upon aggregating FL results, slight variability is observed compared to the centralized counterparts, albeit with comparable mean F1-score (94.77\%) and accuracy (92.30\%) values. Notably, the federated global model exhibits higher specificity (90.65\%) when compared to centralized models such as AE, IF, and OC-SVM, suggesting that FL leverages diverse datasets from individual clients to enhance anomaly detection by minimizing false positives and capturing a broader range of anomalies.

\begin{table}[t]
\setlength\extrarowheight{2pt}
\centering

\begin{tabular}{l|l|l|l|l|l} 
\hline
Metric      & \textbf{Machine} & \begin{tabular}[c]{@{}l@{}} \textbf{Multi}\\ \textbf{func}\\ \textbf{tions}\end{tabular} & \begin{tabular}[c]{@{}l@{}} \textbf{Jaw} \\ \textbf{Crusher}\end{tabular} & \begin{tabular}[c]{@{}l@{}} \textbf{Atlas} \\ \textbf{Copco}\end{tabular} & \textbf{Doosan}          \\ 
\hline
Acc   & min     & 77.54                                                      & 91.65                                                 & 92.43                                                 & 90.95           \\
      & max     & 92.41                                                      & 93.06                                                 & 93.27                                                 & 92.92           \\
      & mean    & \textbf{83.11}                                             & \textbf{92.78}                                        & \textbf{93.00}                                        & \textbf{92.57}  \\
      & std     & 5.76                                                       & 0.40                                                  & 0.21                                                  & 0.51            \\
      & median  & 80.16                                                      & 92.92                                                 & 93.01                                                 & 92.71           \\ 
\hline
Pre   & min     & 82.49                                                      & 95.44                                                 & 97.27                                                 & 95.66           \\
      & max     & 99.26                                                      & 97.81                                                 & 98.07                                                 & 98.07           \\
      & mean    & \textbf{94.70}                                             & \textbf{97.40}                                        & \textbf{97.78}                                        & \textbf{97.10}  \\
      & std     & 6.22                                                       & 0.71                                                  & 0.18                                                  & 0.64            \\
      & median  & 98.14                                                      & 97.72                                                 & 97.78                                                 & 96.92           \\ 
\hline
   & min     & 64.86                                                      & 87.33                                                 & 92.11                                                 & 87.47           \\
TNR & max     & 94.07                                                      & 93.42                                                 & 93.95                                                 & 93.95           \\
      & mean    & \textbf{84.26}                                             & \textbf{92.33}                                        & \textbf{93.31}                                        & \textbf{91.51}  \\
      & std     & 10.24                                                      & 1.82                                                  & 0.45                                                  & 1.70            \\
      & median  & 89.28                                                      & 93.13                                                 & 93.33                                                 & 91.09           \\ 
\hline
  & min     & 76.99                                                      & 92.79                                                 & 91.98                                                 & 91.98           \\
TPR & max     & 94.81                                                      & 93.29                                                 & 93.66                                                 & 93.38           \\
      & mean    & \textbf{85.04}                                             & \textbf{92.94}                                        & \textbf{92.91}                                        & \textbf{92.93}  \\
      & std     & 8.22                                                       & 0.14                                                  & 0.35                                                  & 0.45            \\
      & median  & 78.78                                                      & 92.89                                                 & 92.92                                                 & 93.07           \\ 
\hline
F1    & min     & 86.32                                                      & 94.29                                                 & 94.93                                                 & 93.85           \\
      & max     & 94.80                                                      & 95.32                                                 & 95.43                                                 & 95.22           \\
      & mean    & \textbf{89.14}                                             & \textbf{95.12}                                        & \textbf{95.28}                                        & \textbf{94.97}  \\
      & std     & 3.28                                                       & 0.29                                                  & 0.13                                                  & 0.36            \\
      & median  & 87.46                                                      & 95.22                                                 & 95.29                                                 & 95.05           \\
\hline
\end{tabular}
\caption{Evaluation of local models with Test Dataset}
\label{tab:localmodels}
\end{table}

In assessing the performance of FL models across various clients, evaluating their effectiveness after training on each client is imperative. Notably, higher average results are observed in the Atlas Copco machine, demonstrating an F1-score of 95.28\% and an Accuracy of 93.00\%. Despite its comparatively lower representation, as illustrated in Figure \ref{fig:dataDist_split}, this model exhibits good performance on that machine, underscoring its robustness. Conversely, the latter demonstrates superior results when comparing machines with higher representation in general and test datasets, namely the multifunction machine and the Doosan machine. The Doosan machine yields an average accuracy of 92.57\% versus 83.11\% for the Multifunctions machine, accompanied by an F1-score of 94.97\% compared to 89.14\% and a specificity of 92.93\% versus 85.04\%. These findings suggest a stronger influence of the Doosan machine data on model convergence, potentially attributed to its more prominent representation during weight aggregation, unlike the Multifunctions machine, which may exhibit greater data diversity and lesser representation, impacting later model aggregation.

\begin{table}
\centering
\setlength\extrarowheight{2pt}
\label{tab:Weights}
\begin{tabular}{l|l|l|l} 
\hline
\begin{tabular}[c]{@{}l@{}}\textbf{Hidden Layers per}\\\textbf{Encoder/Decoder}\end{tabular} & \begin{tabular}[c]{@{}l@{}}\textbf{\textbf{Initial}}\\\textbf{\textbf{Layer}}\end{tabular} & \begin{tabular}[c]{@{}l@{}}\textbf{Trainable }\\\textbf{Params}\end{tabular} & \begin{tabular}[c]{@{}l@{}}\textbf{Weights}\\\textbf{(KB)}\end{tabular}  \\ 
\hline
\textbf{1}                                                                                                          & 16                                                                                                                    & 181                                                                          & 0.70                                                                         \\
                                                                                                                    & \textbf{32}                                                                                                           & \textbf{357}                                                                 & \textbf{1.39}                                                                \\
                                                                                                                    & 64                                                                                                                    & 709                                                                          & 2.77                                                                         \\
                                                                                                                    & 128                                                                                                                   & 1143                                                                         & 5.52                                                                         \\ 
\hline
2                                                                                                                   & 16                                                                                                                    & 277                                                                          & 1.08                                                                         \\
                                                                                                                    & 32                                                                                                                    & 805                                                                          & 3.14                                                                         \\
                                                                                                                    & 64                                                                                                                    & 2629                                                                         & 10.26                                                                        \\
                                                                                                                    & 128                                                                                                                   & 9349                                                                         & 36.52                                                                        \\ 
\hline
3                                                                                                                   & 16                                                                                                                    & 293                                                                          & 1.14                                                                         \\
                                                                                                                    & 32                                                                                                                    & 901                                                                          & 3.51                                                                         \\
                                                                                                                    & 64                                                                                                                    & 3077                                                                         & 12.019                                                                       \\
                                                                                                                    & 128                                                                                                                   & 11269                                                                        & 44.019                                                                       \\
\hline
\end{tabular}
\caption{Size of the hidden layer, number of trainable parameters and size of the weights to be sent in each round.}
\label{tab:Weights}
\end{table}

\subsection{Federated learning in LoRaWAN monitoring environments}

This framework was designed for anomaly detection in IIoT-based environments within LoRaWAN networks. In this context, knowing that LoRaWAN networks are characterized by low power, long-range communication, and limited bandwidth, the models’ size becomes a critical consideration. Therefore, for this case study, a particular analysis of the influence of model size is required, as it directly affects the volume of the messages needed to transmit updates in a FL process to complete training on devices. Larger model sizes or larger iterations require more extensive data transmission, which can result in higher energy consumption and longer training times. Consequently, optimizing models’ size or the number of required iterations becomes imperative to ensure the most efficient use of network resources and obtain the best model performance results.

For example, in the optimized architecture shown in Figure \ref{fig:autoencoderStruct}, various sizes of the hidden layer were examined, resulting in different numbers of trainable parameters along with their respective sizes, as shown in Table \ref{tab:Weights}, all this also taking into consideration that they are stored in non-reduced float32 format. Considering the 32 fully connected neurons of the encoder and decoder layers, the combined size of the weights for the updates amounts to \si{1.39}{KB}, which must be transmitted to the server in each round for aggregation. This size roughly corresponds to 357 trainable parameters, encompassing weights and model biases, or with 128 neurons for the hidden layer, potentially reaching \si{5.52}{KB}, which represents a relatively modest weight given the structure. Table \ref{tab:Weights} explores the weights encountered if we add additional hidden layers in the structure, potentially reaching up to \si{44.019}{KB}. However, in the context of low-power networks, even this seemingly modest volume of data transmission could become significant, underscoring the importance of optimizing the model size for efficient communication and operation within these network constraints.

Furthermore, in this subsection, we explore training configurations in FL, aiming to optimize model convergence and performance while considering the constraints of our case study in the construction machinery domain. Initially, we established a baseline using the number of optimized training epochs found in the centralized model, which amounted to 80 epochs. We allocated one epoch for each federated round for our initial comparison, completing 80 rounds. However, recognizing the need to investigate the impact of different combinations of epochs and rounds on model training, we conducted a systematic exploration.

We considered various epoch/round combinations, ranging from 1/80 to 80/1, where the first number denotes the number of epochs per round, and the second number indicates the total number of rounds required to complete the 80 training iterations. Specifically, the explored combinations included 1/80, 2/40, 4/20, 5/16, 8/10, 10/8, 16/5, 20/4, 40/2, and 80/1. By varying the distribution of epochs and rounds while keeping the total training iterations constant, we aimed to discern the optimal configuration that balances communication efficiency and model accuracy within the context of our constrained LoRaWAN network architecture, allowing us to gain insights into the trade-offs associated with different training configurations in FL.

\subsubsection{Estimation of LoRaWAN Message Requirements for Training }

\begin{figure}[t]
\centering
\includegraphics[width=0.99\columnwidth]{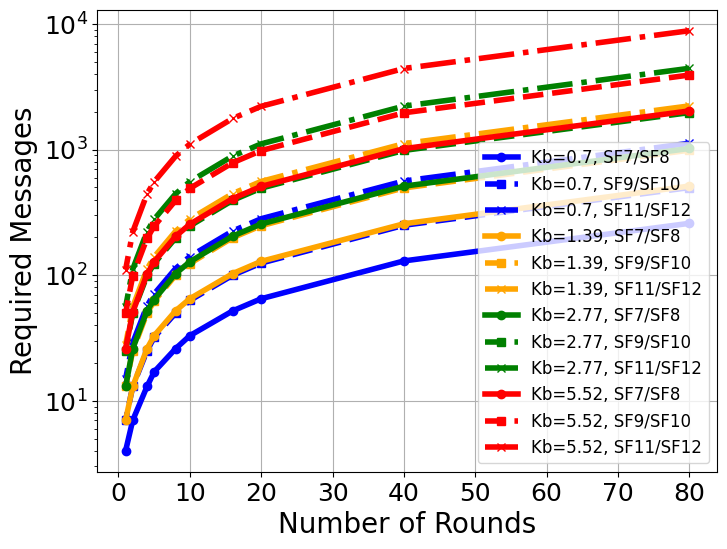}
\caption{Number of LoRaWAN messages needed to complete the training taking the maximum payload size for each SF}
\label{fig:NumberMessages}
\end{figure}

Figure \ref{fig:NumberMessages} provides information on the minimum number of messages required to train LoRaWAN devices within the FL framework. These estimates are calculated based on Equation \ref{ec:NumberofMessages}, where Nm is the number of messages, Msize is the size of the model parameters, MaxPayload is the maximum payload size at each SF without considering the 13-byte header, and Rd is the number of aggregation rounds for FL. Thus, these estimates consider several factors, such as the size of the parameters or weights transmitted during FL rounds, the number of rounds required for convergence, and the payload size con- straints imposed by the different spreading factors (SFs) that can be selected in LoRaWAN networks. In LoRaWAN, the maximum payload capacity varies depending on the SF. SF7 and SF8 support payloads of up to 222 bytes, while SF9 and SF10 are limited to 115 bytes, and SF11 and SF12 can accommodate payloads of 51 bytes, all considering the standard header size of 13 bytes.

\begin{equation}
\text{Nm} = \left\lceil \frac{\text{MSize}}{\text{MaxPayload}} \right\rceil \times \text{Rd}
\label{ec:NumberofMessages}
\end{equation}

Consequently, the number of messages required for training varies significantly depending on the complexity of the model architecture and the chosen SF. The table illustrates that message requirements range from as few as 4 LoRaWAN messages for simpler model structures with a single aggregation round, to 8867 messages for more complex architectures spanning 80 aggregation rounds. For example, considering an optimal model size of \si{1.39}{KB}, the message count can range from 7 to 2233 messages for higher propagation factors, or stabilize at 513 messages for SF7, commonly employed in this type of architecture.

\begin{figure}[t]
\centering
\includegraphics[width=3.5 in]{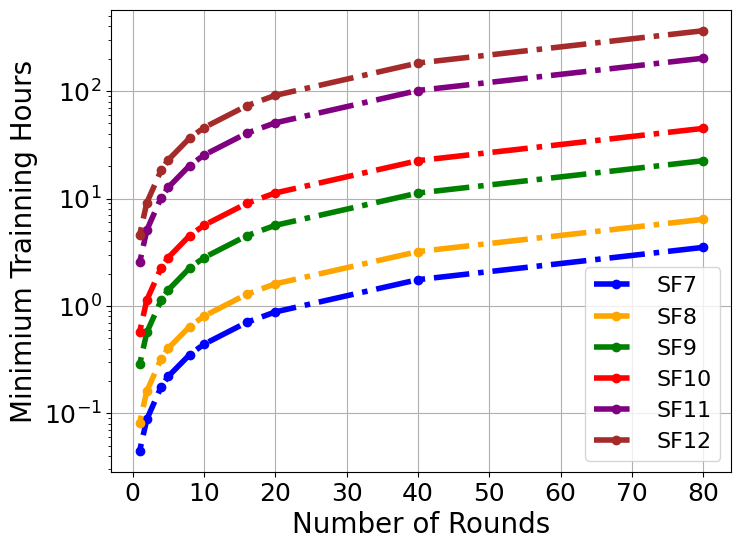}
\caption{Number of training hours for each SF considering as an example 1 hidden layer with size 128}
\label{fig:NumberHours}
\end{figure}

\subsubsection{Considerations for LoRaWAN Communication Constraints}

\begin{figure}[t]
\centering
\includegraphics[width=2.5 in]{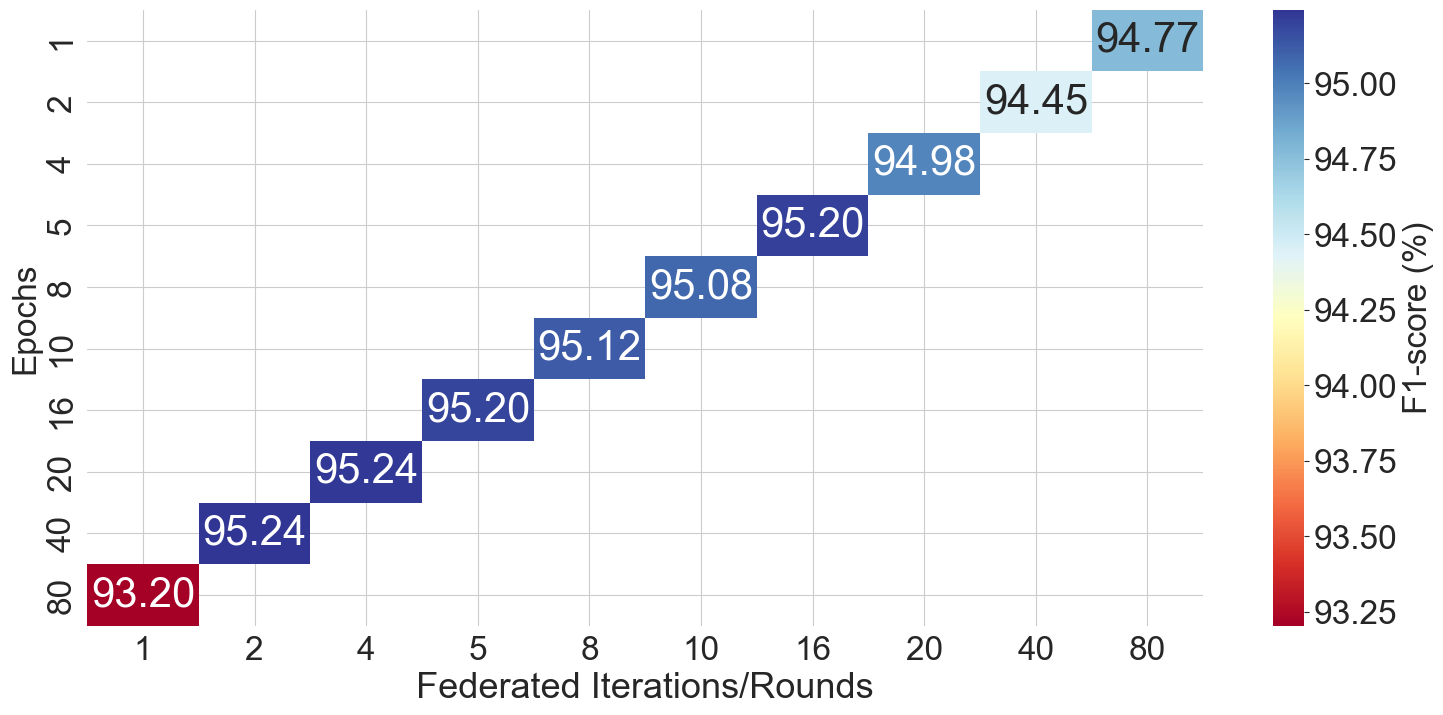}
\caption{F1-score}
\label{fig:F1-scoreep/rd}
\end{figure}

\begin{figure}[t]
\centering
\includegraphics[width=2.5 in]{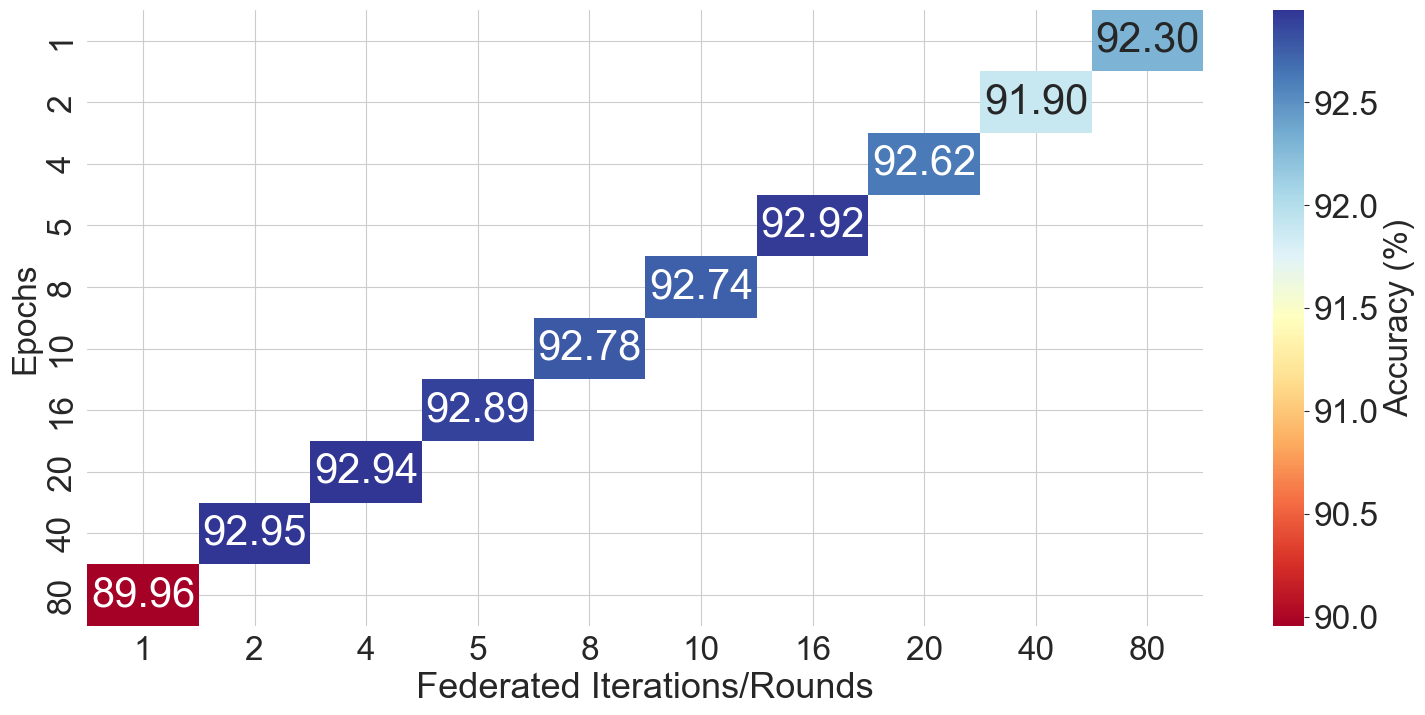}
\caption{Accuracy}
\label{fig:Accuracyep/rd}
\end{figure}

\begin{figure}[t]
\centering
\includegraphics[width=2.5 in]{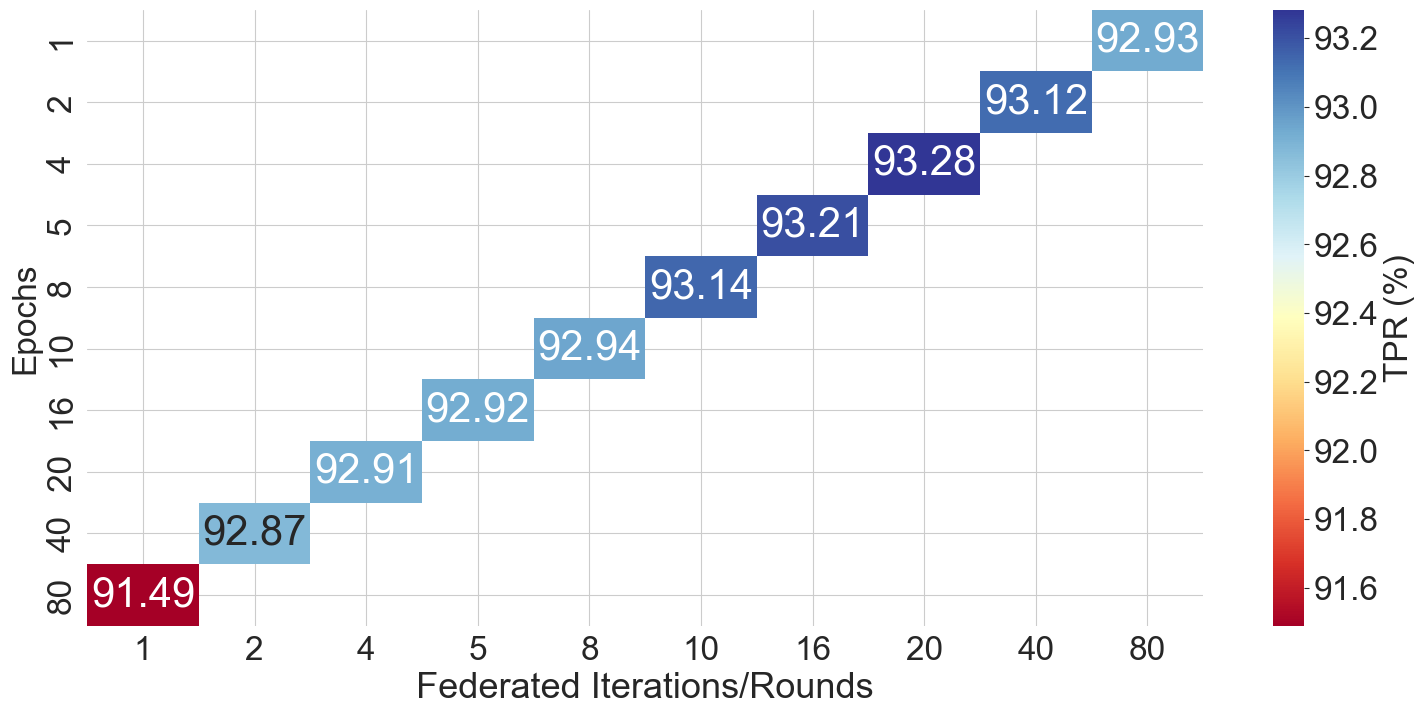}
\caption{TPR}
\label{fig:TPRep/rd}
\end{figure}

\begin{figure}[t]
\centering
\includegraphics[width=2.5 in]{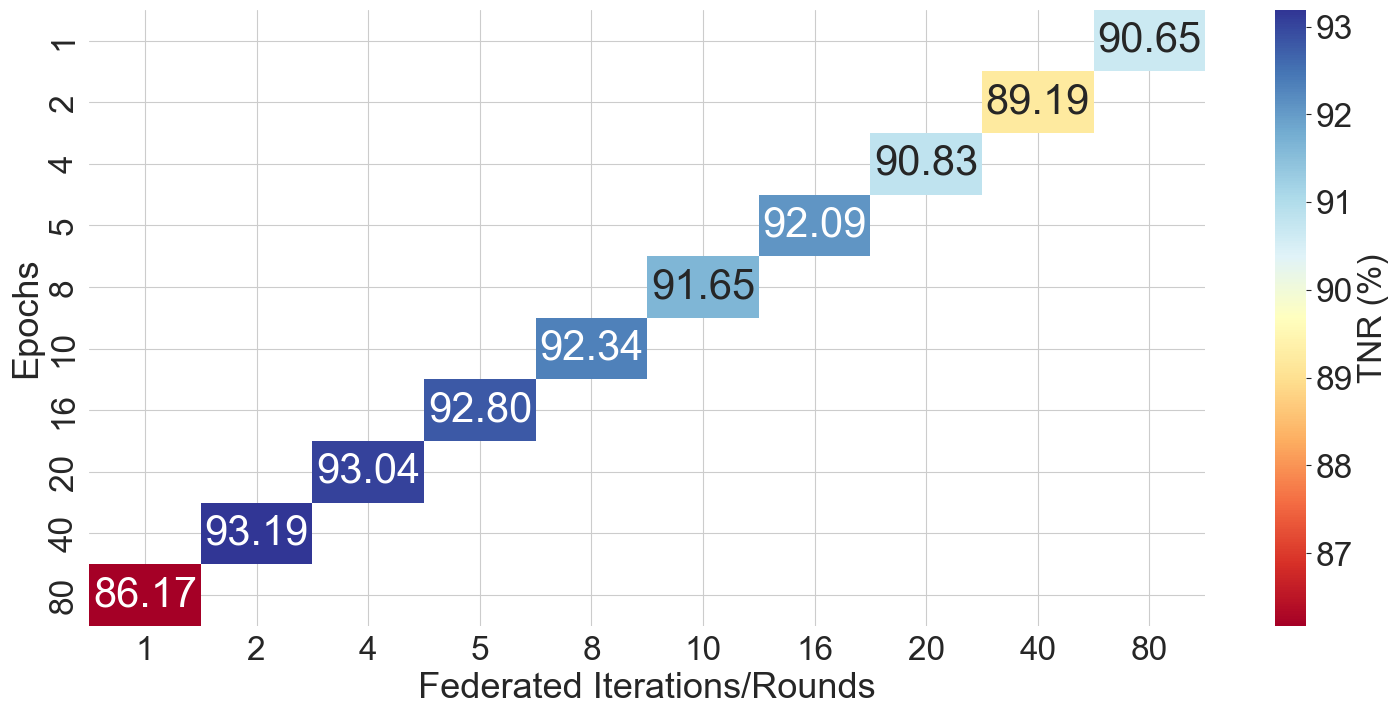}
\caption{TNR}
\label{fig:TNRep/rd}
\end{figure}


by using the ISM band, LoRaWAN imposes limitations not only on the size of the packets per message but also on the transmission time of the messages. These restrictions are also part of Europe’s 868MHz ISM band (similar restrictions exist for other regions), where limitations on message transmission time prevent the constant transmission of LoRaWAN messages and set a cap on the number of messages allowed per SF per hour. Equation \ref{ec:NumberOfHours} allows to calculate the minimum number of hours required to complete the training, where Nh is the number of hours, Nm the number of messages calculated in the previous section, and the minimum periodicity for each SF, where for SF7 it is \si{6.2}{s}, for SF8 it is 11.3 s, for SF9 it is 20.6 s, for SF10 it is 41.2, for SF11 it is 82.3 s, and for SF12 it is 148.3 s. These calculations allow us to estimate the minimum times considering the constraints but do not include the processing time of the models by the devices, which means the times could increase for more complex models. Figure \ref{fig:NumberHours} shows an example of the times achieved for the different rounds and SFs, considering 128 neurons in the hidden layer. The figure shows that the training times could be significantly extended while the number of rounds increases, depending on the modulation factor. As the needed time may be quite long, the importance of optimizing training strategies and model architectures to align with the communication constraints imposed by LoRaWAN networks must be considered and emphasized, ensuring efficient utilization of network resources while facilitating effective FL processes in industrial environments.

\begin{equation}
\text{Nh} =  \frac{\text{Nm} \times \text{MinPeriodicity}}{\text{3600}}
\label{ec:NumberOfHours}
\end{equation}

While the presented calculations do not encompass the time taken by different devices for training nor the processing time for models, they nonetheless provide insight into the minimum time models would require for transmission within Lo- RaWAN architecture structures, considering their constraints. These calculations shed light on the reality of training times within such architectures and underscore potential concerns regarding the duration of training operations. Despite these concerns, it remains imperative to prioritize model updates, leveraging the downtime windows typically available during overnight machine inactivity. Updated models facilitate continuous algorithm refinement, process optimization, and more precise machine performance and operational pattern predictions. Future work will delve into implementing constrained devices, focusing on reducing model size and minimizing message-sending frequency, which aims to streamline training time and resource utilization within LoRaWAN networks.

 \subsubsection{Performance Analysis with Different Epoch and Round Configurations}
Once the estimation of messages and training time was completed, a performance evaluation of the model under various epochs and round configurations was prompted. We computed the average results through multiple independent runs (13) for each configuration. For instance, examining F1-score values in Figure \ref{fig:F1-scoreep/rd}  reveals that better outcomes are obtained when clients execute more epochs per aggregation round, with optimal results observed at 40/20/16 epochs per round. Conversely, the least favorable outcomes occur when employing 80 epochs within a single aggregation round, suggesting that a solitary aggregation for highly trained models could be counterproductive. Similarly, the accuracy trends in Figure \ref{fig:Accuracyep/rd} echo this behavior, indicating better results with 80 rounds and one epoch than 40 rounds and two epochs.
Regarding TNR (Figure \ref{fig:TNRep/rd}) and TPR (Figure \ref{fig:TPRep/rd}), distinct configurations showcase contrasting model performance. Configurations like 2/4/5 rounds exhibit enhanced anomalous value detection with minimized false alarms, strengthening the model's ability to detect anomalies in the machinery, while 16/20/40 rounds demonstrate improved TPR; the parameter is used to evaluate the model's ability to avoid false alarms. Understanding these dynamics is key to optimizing model performance in our case study scenario.

\section{Conclusion and future work}

Integrating intelligence into industrial systems has become essential for enhancing machine operations, particularly in sectors like construction, where maintenance challenges are prevalent. This study focused on implementing an anomaly detection process using data collected within a constrained network architecture like LoRaWAN, targeting the construction industry. Our investigation led to the development of optimized, centralized anomaly detection models, both non-neural-network-based and neural-network-based, with parameters fine-tuned for enhanced performance. For neural-network-based models, we streamlined the structure to reduce size while maximizing efficiency. This involved a two-stage autoencoder optimization process, first minimizing hidden layers and optimizing trainable parameters based on the Mean Squared Error as a loss function. Subsequently, we implemented a second stage to fine-tune the threshold value crucial for anomaly estimation. 
The optimal parameters identified for the centralized autoencoder architecture comprised a single hidden layer with 32 nodes for both the encoder and decoder. The model was trained over 80 epochs with a batch size of 16, employing the Tanh activation function and Adam optimizer. Additionally, a threshold value of 0.1622 was determined for anomaly estimation.
Furthermore, we extended our approach to FL, allowing experiments with clients to iteratively update the model systematically. By leveraging FL, we enabled decentralized training while maintaining model accuracy and relevance across diverse datasets. We performed tests with the optimized structure of the centralized autoencoder, executing FL with four clients, each representing one of the study machines. Utilizing an optimized neural network structure yielded results comparable to those of its centralized counterparts. We achieved an average accuracy rate of 92.30\% and an F1-score of 94.77, showcasing the model’s efficacy.

Moreover, the neural network structure demonstrated enhanced capability in mitigating false alarms, as evidenced by the TPR parameter, which surpassed 90\%. Furthermore, the model exhibited robust performance across individual machines, with accuracy ranging from 83.11\% for the Multifunctions machine to a maximum of 93\% for the other machines. F1 scores ranged from 89.14 for the Multifunctions machine to 95.28 for others, underscoring the influence of data diversity on performance.
Finally, we addressed practical considerations such as the minimum number of messages required for training over LoRaWAN and estimated training times, considering airtime constraints in the ISM band used. We evaluated the use of different configurations with round/epoch balancing of the models. We found that the optimal configuration consisted of 20 epochs and four rounds, leading to favorable results like an F1- score of 95.23, an accuracy of 92.94, and a TNR of 93. Moreover, we found that it is essential to consider the potential drawbacks of using fewer aggregation rounds in FL, especially in constrained devices and LoRaWAN networks. While fewer aggregation rounds may seem advantageous due to reduced message transmission and shorter training times, they can pose significant challenges. One concern is the risk of incomplete or less globally representative models. With fewer aggregation rounds, models may not adequately capitalize on the diverse display launched across devices, leading to suboptimal global model performance.
Moreover, fewer aggregation rounds may exacerbate convergence issues, potentially resulting in the instability of the global model. On the other hand, increasing the number of epochs on the devices to compensate for the lower number of aggregation rounds may also lead to difficulties. A higher number of epochs places a higher computational load on the devices, which may prolong training execution time and overload device resources. Finding a balance between training complexity and the time saved by reducing aggregation rounds is crucial. In addition, an excessive number of epochs can saturate device resources and lead to training failures. Implementing recovery algorithms to solve these problems adds complexity to the system and may affect the overall performance. Therefore, future research efforts will delve deeper into these complex issues to address the challenges of FL on constrained devices over LoRaWAN networks. By exploring strategies to mitigate the limitations of fewer aggregation rounds and optimizing training configurations for resource-constrained environments, we aim to improve the effectiveness and efficiency of FL in real-world applications. In addition, the performance of the models should be analyzed by considering size reduction or TinyML techniques to reduce complexity and be compatible with resource-constrained devices such as microcontrollers.
\label{sec:Conclusion}

\bibliographystyle{IEEEtran}
\bibliography{library.bib}

\vspace{11pt}


\vspace{-30pt}

\vfill

\end{document}